\documentclass[letterpaper, 10 pt, conference]{ieeeconf}  

\IEEEoverridecommandlockouts                              

\usepackage{hyperref}
\usepackage{amsmath}
\usepackage{amssymb}
\usepackage{booktabs}
\usepackage{graphicx}
\usepackage{siunitx}
\usepackage{tabularx}
\usepackage{array}
\usepackage{subcaption}
\usepackage{makecell}
\usepackage{multirow}
\usepackage{stfloats}
\newcolumntype{Y}{>{\centering\arraybackslash}X}

\title{\LARGE \bf
Robotic Grasping and Placement Controlled by EEG-Based Hybrid Visual and Motor Imagery
}

\author{%
Yichang Liu$^{1}$,
Tianyu Wang$^{2,5}$,
Ziyi Ye$^{3,4}$\textsuperscript{*},
Yawei Li$^{6}$,
Yu-Gang Jiang$^{3,4}$,
Shouyan Wang$^{1}$\textsuperscript{*},
Yanwei Fu$^{2,4,5}$%
\thanks{This work was supported by STI 2030 Major Projects (2021ZD0200407),
the National Natural Science Foundation of China (62521004),
the National Key Research and Development Program of China (2022YFC2405100), the Shanghai Pujiang Talents Program (Grant 24PJD006), and the AI for Science Foundation of Fudan University.}%
\thanks{Project Page: \href{https://yichangliu1224.github.io/EEG-Graspbot/}{https://yichangliu1224.github.io/EEG-Graspbot/}}%
\thanks{All authors are affiliated with
$^{1}$Institute of Science and Technology for Brain-inspired Intelligence (ISTBI), Fudan University;
$^{2}$School of Data Science, Fudan University;
$^{3}$Institute of Trustworthy Embodied AI, Fudan University;
$^{4}$Shanghai Key Laboratory of Multimodal Embodied AI;
$^{5}$Shanghai Innovation Institute;
$^{6}$ETH Z\"urich.}%
\thanks{\textsuperscript{*} indicates the corresponding authors}%
}

\begin{document}

\IEEEpeerreviewmaketitle
\maketitle

\thispagestyle{empty}
\pagestyle{empty}


\begin{abstract}

We present a framework that integrates EEG-based visual and motor imagery (VI/MI) with robotic control to enable real-time, intention-driven grasping and placement. Motivated by the promise of BCI-driven robotics to enhance human-robot interaction, this system bridges neural signals with physical control by deploying offline-pretrained decoders in a zero-shot manner within an online streaming pipeline. 
This establishes a dual-channel intent interface that translates visual intent into robotic actions, with VI identifying objects for grasping and MI determining placement poses, enabling intuitive control over both what to grasp and where to place. The system operates solely on EEG via a cue-free imagery protocol, achieving integration and online validation. 
Implemented on a Base robotic platform and evaluated across diverse scenarios, including occluded targets or varying participant postures, the system achieves online decoding accuracies of 40.23\% (VI) and 62.59\% (MI), with an end-to-end task success rate of 20.88\%. These results demonstrate that high-level visual cognition can be decoded in real time and translated into executable robot commands, bridging the gap between neural signals and physical interaction, and validating the flexibility of a purely imagery-based BCI paradigm for practical human–robot collaboration.

\end{abstract}
\section{INTRODUCTION}

Recently, robotic technologies have achieved remarkable progress in end-to-end grasping and manipulation~\cite{zhang2022robotic, kleeberger2020survey, billard2019trends}. Modern robots, empowered by deep learning and advanced control algorithms, can recognize objects, plan complex motions, and interact with real-world environments~\cite{wang2024polaris, shen2023distilled, bjorck2025gr00t}. 
However, they still struggle to interpret and execute commands derived from human \textit{high-level} intentions~\cite{zhang2023noir, zhang2025mind}, such as natural language instructions, context-dependent requests, and multimodal cues. In contrast, over the past two decades, \textit{low-level} robotic capabilities have matured significantly, enabling reliable perception, motion, object manipulation, and basic action execution.
Consequently, bridging these robust low-level skills with high-level cognitive processes that reflect human intent remains a central challenge.



%

To address this challenge, studies have explored combining cognitive signals with robotic control. Early approaches used explicit cues, such as voice commands, hand gestures, or eye tracking, to infer user intent and guide robot actions~\cite{cheang2022learning, shafti2019gaze}. More recently, brain-computer interfaces (BCIs) have been integrated directly, allowing robots to interpret a user’s brain activity related to object recognition or movement intentions, without requiring physical or verbal input.

However, the integration of brain signals with robotic actions remains in its early stages and faces significant challenges.
Most existing BCI-robot control systems rely on constrained and explicit control signals such as event-related potentials (e.g., P300), steady-state visually evoked potentials (SSVEP)~\cite{beverina2003user}, or motor imagery (MI)~\cite{leeb2007brain}. 
While effective in laboratory environments, these methods are often restricted to predefined commands or simple actions (e.g., ``move left'' or ``move right''), making them ill-suited for generating complex, intuitive, and natural-intention-driven commands for real-world physical interaction. 
This gap between a user’s cognitive intent and the robot’s executable actions remains a major challenge for the widespread adoption of BCI technology~\cite{varbu2022past, tai2024brain}.


Recent advances in brain decoding, particularly with electroencephalography (EEG), have increasingly targeted higher-order cognitive states such as visual perception and imagery~\cite{wilson2024feasibility, kilmarx2024evaluating, gao2025cinebrain}. EEG, as a non-invasive and portable modality, shows promise in interpreting complex brain activity. 
Nevertheless, most studies have focused on passive observation and offline analysis, without directly translating these rich cognitive signals into real-time robotic actions. As a result, the potential of high-level visual cognition for controlling robotic systems remains largely underexplored.




\begin{figure*}[!t]
    \centering
    \includegraphics[width=1.0\linewidth]{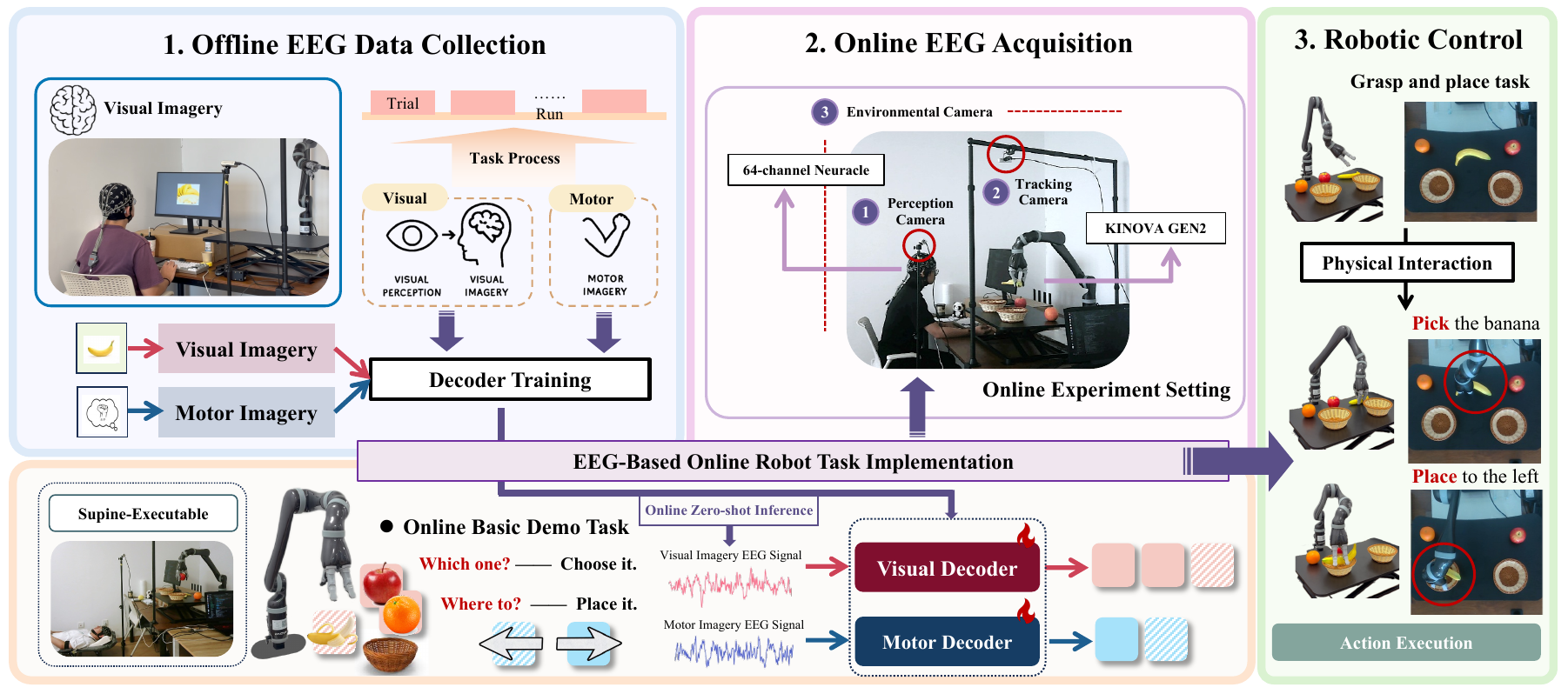}
    \caption{\textbf{High-level cognitive control pipeline for EEG-based robotic manipulation.}
    The framework consists of three main components:
    (1) \textbf{Offline EEG data collection}: visual imagery (VI) and motor imagery (MI) EEG data are collected and labeled to train visual and motor decoders.
    (2) \textbf{Online EEG acquisition}: A dual-channel system is deployed in which VI-EEG is decoded into grasping intentions and MI-EEG determines placement positions, enabling real-time task-level command generation.
    (3) \textbf{Robotic control}: the trained decoders drive a robotic arm to perform the grasp and place task in real-world, demonstrating the seamless mapping from high-level visual cognition to physical manipulation.    \label{fig:framework} }
    \vspace{-0.15in}

\end{figure*}

In this study, we present an end-to-end framework that integrates high-level visual cognition with robotic control for a pick-and-place task. 
As shown in Figure~\ref{fig:framework}, our \textbf{EEG-based cognitive control pipeline} first collects both visual imagery (VI) and motor imagery (MI) EEG data from five participants in an offline setup to pretrain the corresponding decoders. The selected decoders are then integrated into the robotic control system, where VI selects the target object for grasping and MI specifies the placement pose. 
The offline evaluation achieves average accuracies of 44.11\% for VI and 76.53\% for MI, respectively.
Motivated by the offline results, 
we deployed the VI/MI decoders in a zero-shot online setting on a dual-channel EEG system for a pick-and-place platform, achieving online accuracies of 40.23\% (VI) and 62.59\% (MI) with an end-to-end task success rate of 20.88\%. 
This \textbf{online application} demonstrates that VI and MI can convey \textbf{higher-level intentions} directly usable for task execution, and the system remains effective in scenarios such as control in supine position without direct sight of the physical objects.

Our main contributions are:
\begin{enumerate}

\item We propose and empirically validate a high-level cognitive control paradigm as a natural-intention-driven BCI grounded in \textbf{visual imagery (VI)}, translating VI-EEG signals directly into task-level robotic commands for pick-and-place manipulation.

\item We develop a \textbf{real-time framework} that bridges brain signals and robotic actions via a dual-channel intent interface: VI-EEG decodes object intent and MI-EEG infers target poses, achieving a seamless mapping from high-level visual cognition to physical manipulation.

\item We perform \textbf{end-to-end system integration} and conduct online system validation, implementing functional demonstration on a robotic platform, which achieves precise real-world grasping and placement, and showcases advanced interaction capabilities. 

\end{enumerate}


\section{RELATED WORKS}

\subsection{Visual Decoding from Brain Signals}

Visual decoding from brain signals has become a rapidly advancing field, with significant strides made in translating neural activity into interpretable visual information\cite{miyawaki2008visual, lan2023seeing}. Early work relied on fMRI due to its high spatial resolution, enabling reconstruction of simple and natural images from voxel patterns, and coupled with the recent integration of latent diffusion models improving semantic fidelity~\cite{huo2024neuropictor, gao2025mind, gao2025cinebrain}. However, its inherent cost, immobility, and low temporal resolution limit its suitability for real-time BCI~\cite{yang2025applications}. In contrast, EEG/MEG offer unparalleled temporal resolution in the millisecond range, with EEG being particularly suitable for real-time and online BCI systems~\cite{he2020brain}. Over the years, EEG-based decoding methods have evolved from basic CNNs and temporal networks to more complex architectures~\cite{li2025deep}, including attention-based and graph-augmented models~\cite{bagchi2022eeg, kalafatovich2022learning}. Some models also incorporate alignment with large-scale vision and multimodal datasets~\cite{du2023decoding,song2023decoding}. However, most existing studies primarily focus on offline classification and passive observation~\cite{guo2024eegvision, li2024visual}, often failing to connect decoded visual semantics to actionable tasks in real-time scenarios. As a result, bridging high-level visual imagery with actionable robotic commands remains an ongoing challenge.

\subsection{Robotics Control using EEG}

Prevailing BCI-robot paradigms typically rely on P300, SSVEP, or MI signals. While these methods are reliable in controlled settings, they are limited by predefined commands and often require the user to focus on flickering stimuli~\cite{abiri2019comprehensive, kim2023eegssvep, hameed2025enhancing}, which diminishes naturalness and can lead to user fatigue. Consequently, these approaches struggle to facilitate intuitive, task-level interactions in more complex and dynamic environments. 
Even in more naturalistic settings, grounding implicit \emph{high-level} intent into physically feasible robot actions remains challenging~\cite{zhang2023noir, zhang2025mind}. This motivates more direct, action-oriented intent interfaces.
For real-time BCI applications, approaches such as LDA, SVMs, and shallow neural networks are often preferred due to their simplicity~\cite{dos2023comparison, carrara2025geometric}. However, even more advanced techniques, including pretrained neural networks, are typically restricted to offline processing. Therefore, decoding zero-shot visual imagery EEG into robotic grasping and placement actions is vital for bridging the gap between laboratory-based BCI systems and real-world applications, allowing for more intuitive and natural interactions in diverse environments.

\section{DATA COLLECTION AND SYSTEM CONFIGURATION}

\subsection{Participants} Five healthy graduate students~(2 females; age 23–30 years; M=25.4, SD=2.97) from a university participated in this study. 
All were right-handed with normal or corrected-to-normal vision and hearing, and none reported neurological disorders. Two participants had prior experience with EEG experiments. Written informed consent was obtained and the protocol was approved by the appropriate ethics committee\footnote{Ethics approval was granted by the Ethics Committee of Fudan University, approval number FE25500I.}. Each participant received approximately \textdollar42 USD per experimental session. All five participants completed the offline experiments, and four further engaged in the online testings.

\subsection{Task Paradigms} 
Offline data collection comprised visual perception~(VP), visual imagery~(VI), and motor imagery~(MI) tasks. 
This study involved a single offline session, with continuous EEG signal recording throughout the experiment. 

\textbf{\textit{Visual Perception (VP)\& Visual Imagery(VI):}}
Stimuli were drawn from the public \textit{Fruit Images for Object Detection} dataset\footnote{Aliyun Tianchi: \href{https://tianchi.aliyun.com/dataset/93848}{https://tianchi.aliyun.com/dataset/93848}}, including three easily recognizable fruit categories (apple, banana, and orange).
Each trial began with a 1-second fixation period marked by a central cross, followed by a 2-second presentation of the target image (perception phase). 
Participants were then instructed to press the space bar to perform eyes-closed visual imagery of the presented object for 5 seconds (imagery phase), a duration supported by previous studies on visual imagery timing~\cite{kilmarx2024evaluating, lee2022exploring}.
The imagery period ended with a 300~ms auditory tone (PsychoPy default pitch ``A'', approximately 440~Hz). 
After each trial, participants reported their attentional engagement on a 10-point scale (0 = not at all focused, 9 = fully focused). 
The collected average attention level for each subject in each trial of the VI task was 8.25 $\pm$ 1.02 (range: 0-9), indicating a high attention during the study.

The VI task comprised 10 runs, each including 3 blocks of 10 randomized trials. 
Short breaks (10 sec) were provided between blocks, and participants rested for 1--2 min between runs. Each run lasted approximately 6--7 min (30 trials), yielding 300 trials in total ($\sim$90 min).

\textbf{\textit{Motor Imagery (MI):}}
The MI protocol followed the BCI Competition IV-2b design~\cite{leeb2007brain}. 
Each trial started with a 1-second fixation period with the center cross, followed by a 1.25-second directional arrow cue (left or right). A short auditory warning tone (70 ms, approximately 440 Hz) was presented 1 second after cue onset to prompt imagery initiation. Participants then performed kinesthetic motor imagery of the cued hand for 4 seconds, and a 200~ms auditory tone signaled the end of the trial. Each trial was followed by a 1.5-second pause and a 1.5-second ITI, which allowed the EEG signal to recover and reduced temporal overlap between consecutive trials.

Each run consisted of 20 trials with balanced, randomly intermixed left/right cues. Participants completed 5 runs (100 trials in total), with each run lasting approximately 3 minutes. The entire MI session lasted 20–30 minutes.

\begin{figure}[h!]
\vspace{-0.1in}
    \centering
    \includegraphics[width=0.9\columnwidth]{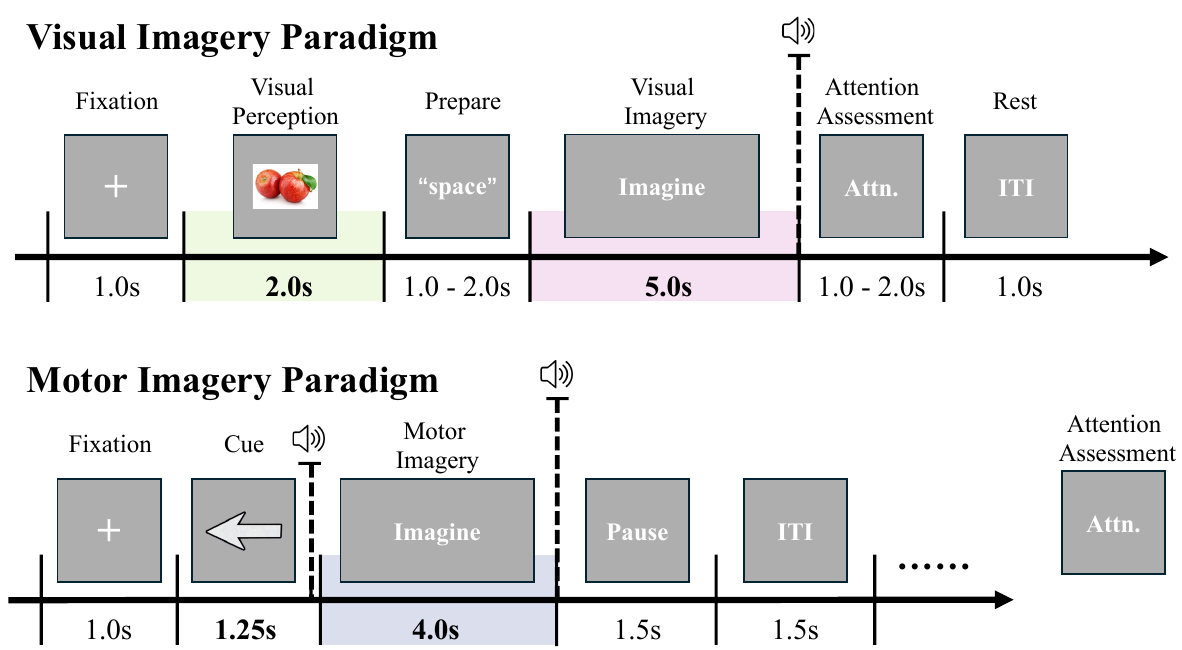} 
    \caption{Visual Imagery and Motor Imagery Paradigms.}
    \label{fig:paradigm}
\vspace{-0.12in}
\end{figure}

The visual imagery and motor imagery paradigms are shown in Figure~\ref{fig:paradigm}. For each task, the phase triggers were recorded by PsychoPy and transmitted to the EEG system, marking the start and end of every phase. The total duration of all offline sessions was approximately 2 hours.

\subsection{Signal Acquisition and Preprocessing}
EEG was recorded with a 64-channel Neuracle system~(59 scalp EEG, 2 mastoid, 2 EOG, 1 ECG, 1000 Hz sampling; 24-bit A/D resolution; $\pm$0.003\% accuracy; analog input range $\pm$5 V) on the NeuroHUB platform, keeping electrode impedances below 10~k$\Omega$. 
Stimuli were presented on a 24.5-inch, 100~Hz monitor at a 50--70 cm viewing distance.
Preprocessing was performed in MNE-Python, including standard procedures such as band-pass filtering, re-referencing to linked mastoids~(HEOR and HEOL), artifact removal using Independent Component Analysis~(ICA), and artifact identification based on correlations with recorded signals~(threshold $>$ 0.95).
Three band-pass versions were created for comparison (0.5--40 Hz, 0.5--60 Hz, 0.5--100 Hz), and a 50 Hz notch filter was applied to suppress line noise. 
After preprocessing, we collect two sets of offline data (VI and MI) for each participant, with three band-pass filtered versions to enable the exploration of different high-frequency (e.g., high gamma) components in subsequent analyses. 

\subsection{Robotics Platform}

We employed a KINOVA GEN2 robot as the embodied platform for systematic interaction. Two Intel RealSense D435 cameras served as the primary sensors of the robotic perception system. The camera positioned directly in front of the workstation acted as the main sensor and was extrinsically calibrated in an eye-to-hand setup to support both the perception system and model inference. An additional camera, mounted vertically above the workstation, provided a top-down view during online inference, enabling execution monitoring and tracking of object interaction. Furthermore, we installed an XWF-1080P6 RGB camera outside the workstation to capture global information for the EEG-based robotic grasping system.

\section{METHODOLOGY AND EVALUATION}



Prior to deploying the online system, we trained and validated our EEG decoding models using offline datasets to establish a performance baseline.

\subsection{Model Selection}
This study refers to the model selections and experimental setups in EEG-ImageNet~\cite{zhu2024eeg} with specific modifications tailored to the task of decoding distinct intentions for real-time robotic control driven by brain signals.
The models encompass both traditional machine learning algorithms, including support vector machine (SVM), random forest, K-nearest neighbors (KNN), decision tree (DT), and ridge regression, as well as deep learning models such as multi-layer perceptron (MLP), EEGNet~\cite{lawhern2018eegnet}, and RGNN~\cite{zhong2020eeg}. 
MLP is a fundamental feedforward neural network, while EEGNet is a lightweight convolutional neural network specifically designed for EEG signals, and RGNN is a recurrent graph neural network that effectively captures the spatio-temporal dependencies within EEG data.

\subsection{Feature Extraction}

Considering the cognitive processes involved in visual and motor intention processing, we removed 10\% of the signal from both the beginning and end of each EEG segment to ensure the neural signals were entirely confined to the task-relevant period. 
The remaining signal segments were split into 500 ms sliding windows with minimal overlap.
The segmented signals were used as features for EEGNet which requires time-domain signals as input. 
For other models, \textbf{Differential Entropy (DE)}~\cite{wang2021review} are extracted as the input feature, which estimate $P(f)$ for each frequency band using the following formula:
\begin{equation}
    P(f) = \lim_{T \to \infty} \frac{1}{T} |X_T(f)|^2,
\end{equation}

where $f$ denotes the (continuous) frequency (Hz), $P(f)$ represents the Power Spectral Density (PSD) at $f$, and $X_T(f)$ is the Fourier transform of the truncated signal $x(t)$ over the interval $[0,T]$. Then, DE is computed as


\begin{equation}
    DE = -\int_{\mathcal{F}} P(f)\,\log\!\bigl(P(f)\bigr)\, df,
\end{equation}

where $\mathcal{F}$ is the frequency range 
of interest.


In addition to the traditional frequency bands ($\delta$ (0.5-4 Hz), $\theta$ (4-8 Hz), $\alpha$ (8-13 Hz), and $\beta$ (13-30 Hz)), we specifically computed three new $\gamma$ bands: (30-40 Hz), (30-60 Hz), and (30-100 Hz), tailored to different filtering strategies. 
This customization was designed to more effectively capture individual brain activity states for intention decoding. 
For each EEG signal segment, we obtained differential entropy values for each electrode and frequency band.


\subsection{Experimental Setup}

The MLP was implemented with two hidden layers, utilizing the ReLU activation function. 
Both the EEGNet and RGNN models were implemented strictly according to the original architectures. 
To ensure model generalizability and accurate evaluation, each participant's dataset was partitioned using stratified sampling into an 80\% training set and a 20\% test set, which maintained a consistent class label distribution. 
Experiments were conducted on a server equipped with an NVIDIA A6000 GPU, leveraging Python 3.11.11 and PyTorch 2.6.0. 
Each model was trained for 1000 epochs.

\subsection{Experimental Results}

\paragraph{Offline Performance}

\begin{table}[t]
    \centering
    \caption{\textbf{Offline classification results for different models in VP, VI and MI tasks.} The mean decoding accuracy across all subjects is reported for each model in different frequency conditions across three tasks. \textbf{Bolded values} indicate the highest accuracy in each row, while \underline{underlined values} denote the second highest accuracy to report competitive alternatives for model selection.
    \label{tab:offline_results}}
    \begin{tabular*}{\columnwidth}{@{\extracolsep{\fill}}c ccccc}
        \toprule
        \textbf{Subject} & \textbf{EEGNet} & \textbf{MLP} & \textbf{RGNN} & \textbf{SVM} & \textbf{DT} \\
        \midrule
        \multicolumn{6}{c}{\textbf{Visual Perception}} \\
        \midrule
        00 & \textbf{0.4844} & \underline{0.4767} & 0.4622 & 0.4067  & 0.4033  \\
        01 & \textbf{0.4911}  & \underline{0.4667}  & 0.4278  & 0.4211  & 0.3711  \\
        02 & \textbf{0.4678}  & \underline{0.4522} & 0.4155  & 0.3833  & 0.3667  \\
        03 & \textbf{0.4978}  & \underline{0.4533}  & 0.4322  & 0.4011  & 0.3911  \\
        04 & \textbf{0.4911} & \underline{0.4889} & 0.4633 & 0.4633 & 0.3978 \\
        \cmidrule(lr){1-6}
        \textbf{Average} & \textbf{0.4864}  & \underline{0.4676}  & 0.4402  & 0.4151  & 0.3860 \\
        \midrule
        \multicolumn{6}{c}{\textbf{Visual Imagery}} \\
        \midrule
        00 & 0.3979 & \textbf{0.4702} & \underline{0.4451} & 0.4014 & 0.3694 \\
        01 & 0.3688 & \textbf{0.4069} & \underline{0.3903} & 0.3694 & 0.3549 \\
        02 & 0.3639  & \textbf{0.4173}  & \underline{0.3868}  & 0.3639  & 0.3451  \\
        03 & 0.3736  & \textbf{0.4757}  & \underline{0.4507}  & 0.3826  & 0.4111  \\
        04 & 0.3694  & \textbf{0.4354}  & \underline{0.4299}  & 0.4028  & 0.3521  \\
        \cmidrule(lr){1-6}
        \textbf{Average} & 0.3747  & \textbf{0.4411}  & \underline{0.4206} & 0.3840  & 0.3665  \\
        \midrule
        \multicolumn{6}{c}{\textbf{Motor Imagery}} \\
        \midrule
        00 & \textbf{0.7967}  & \underline{0.6983}  & 0.6567  & 0.5850  & 0.5250   \\
        01 & \textbf{0.6933} & \underline{0.6783} & 0.6567  & 0.6317  & 0.5550 \\
        02 & \underline{0.7250}  & \textbf{0.7267} & 0.7117 & 0.6867 & 0.5933 \\
        03 & \textbf{0.7850} & \underline{0.6917} & 0.6700 & 0.6183 & 0.5850 \\
        04 & \textbf{0.8267} & \underline{0.8167} & 0.7800 & 0.7017 & 0.6450 \\
        \cmidrule(lr){1-6}
        \textbf{Average} & \textbf{0.7653} & \underline{0.7223} & 0.6950 & 0.6447 & 0.5807 \\
        \bottomrule
    \end{tabular*}
\vspace{-0.2in}
\end{table}

\begin{figure*}[!htbp]
    \centering
    \begin{subfigure}[t]{0.32\textwidth}
        \centering
        \includegraphics[width=\linewidth]{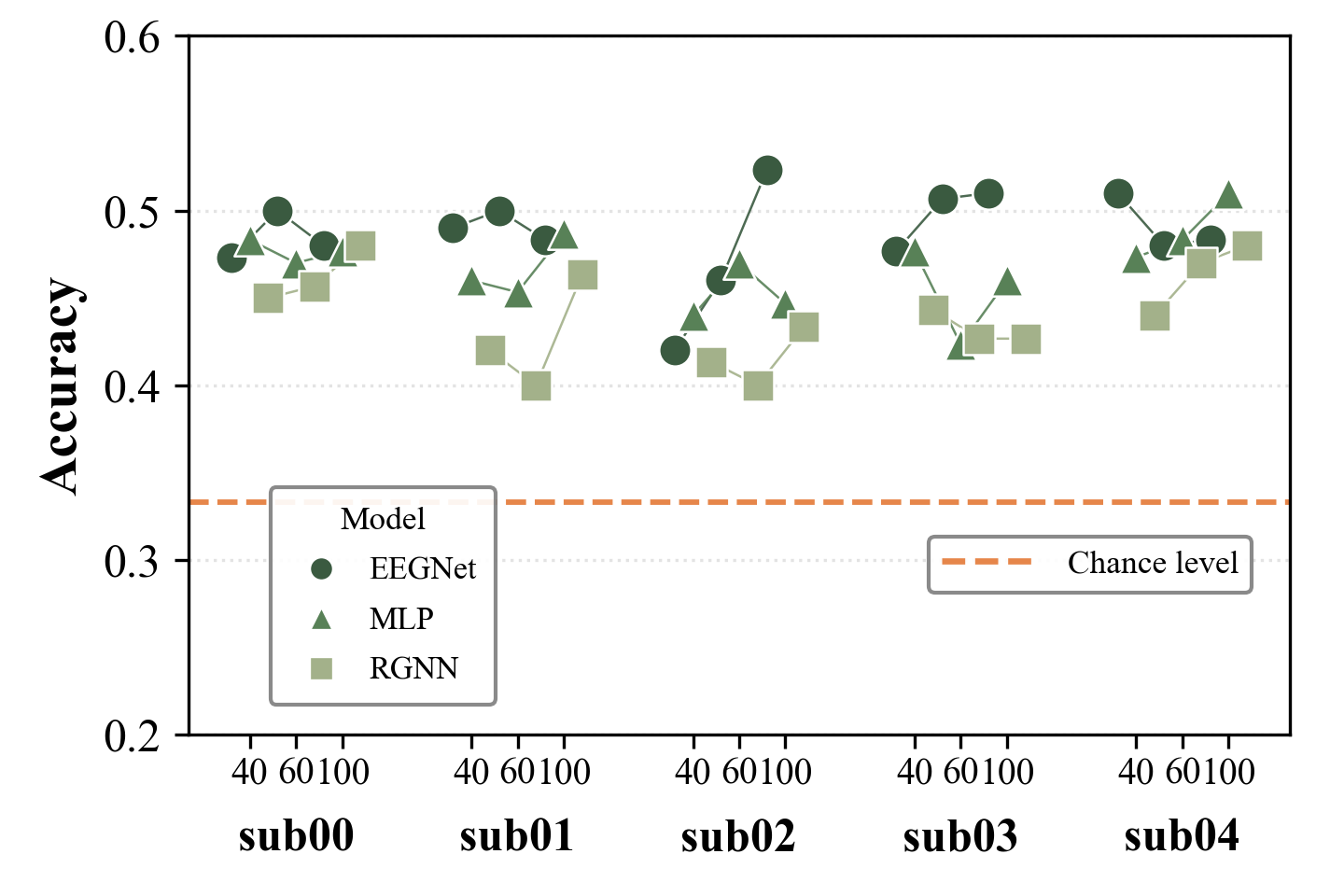}
        \caption{Visual Perception}
        \label{fig:offline:perception}
    \end{subfigure}
    \hfill
    \begin{subfigure}[t]{0.32\textwidth}
        \centering
        \includegraphics[width=\linewidth]{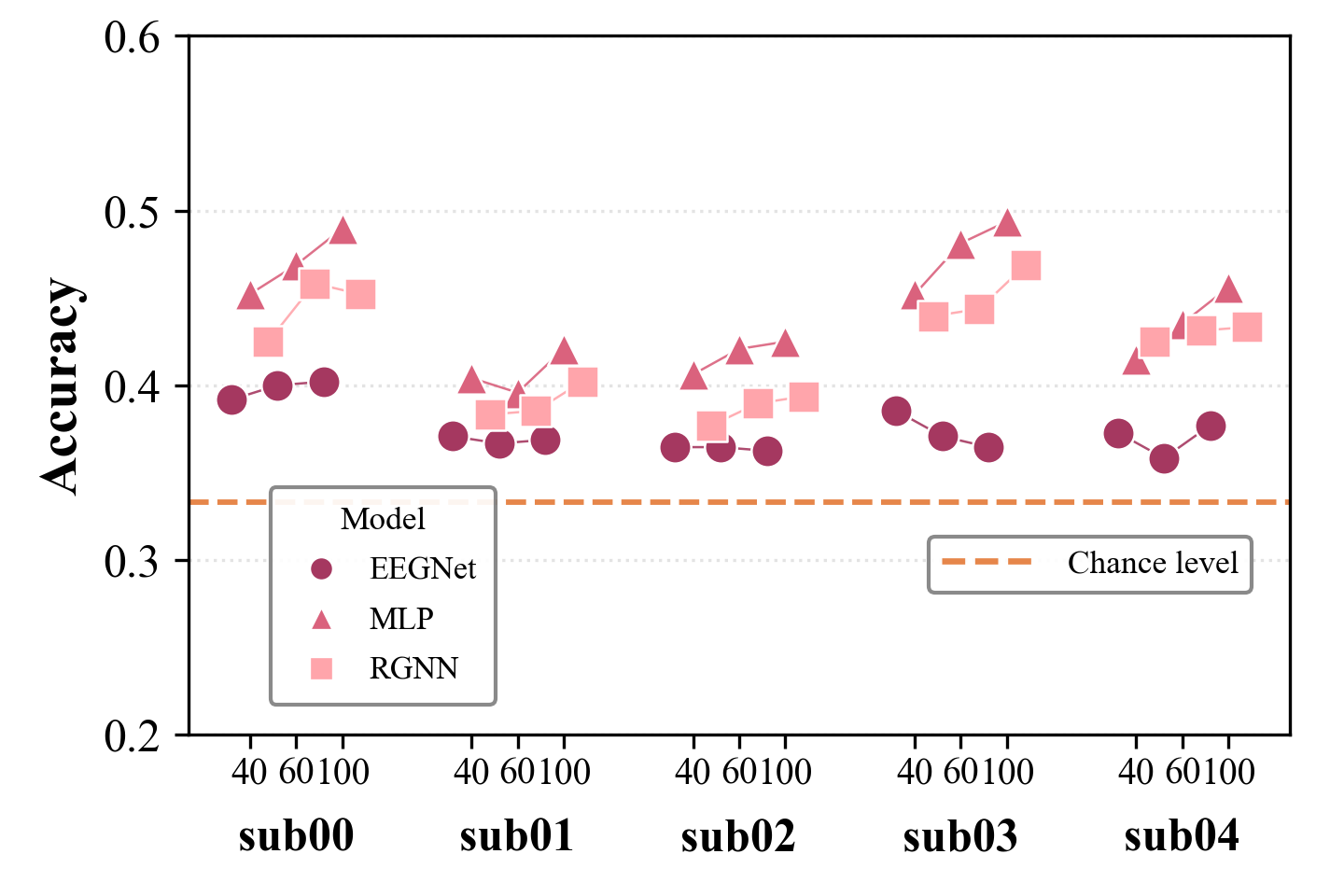}
        \caption{Visual Imagery}
        \label{fig:offline:imagery}
    \end{subfigure}
    \hfill
    \begin{subfigure}[t]{0.32\textwidth}
        \centering
        \includegraphics[width=\linewidth]{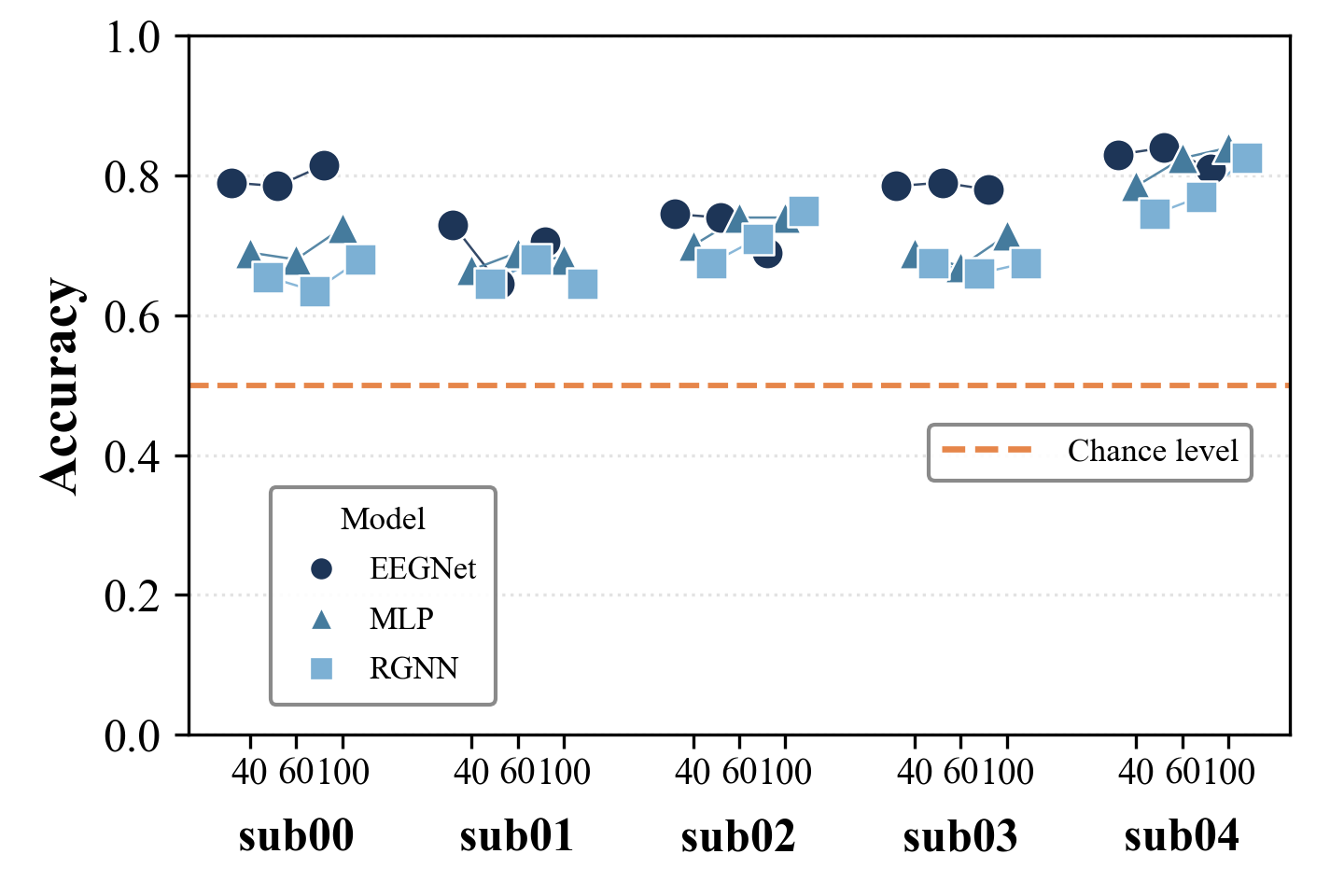}
        \caption{Motor Imagery}
        \label{fig:offline:motor}
    \end{subfigure}

    \caption{\textbf{Offline results for three tasks: (a) Visual Perception, (b) Visual Imagery, and (c) Motor Imagery.} 
    Points show per-subject accuracies across models and frequency settings; the dashed line denotes chance level (VP/VI: 33.3\%, MI: 50\%). MI consistently outperforms VP/VI, and all tasks exceed chance.
    \label{fig:offline}}
\vspace{-0.2in}
\end{figure*}

The offline data results for five subjects under different models are summarized in Table~\ref{tab:offline_results} and Figure~\ref{fig:offline}, representing the mean decoding accuracy across all subjects. 
The MLP model achieved the highest average accuracy in the offline-VI task, with an accuracy of 44.11\%, while EEGNet achieved an average accuracy of 76.53\% in the offline-MI task. 
This performance disparity can be attributed to the fundamental differences in brain signal generation. 
Motor imagery tasks are known to elicit strong and distinct event-related desynchronization/synchronization (ERD/ERS) patterns in the $\mu$ and $\beta$ frequency bands, which are more readily detectable and classifiable from EEG signals~\cite{yang2025multi, zhu2025study}. 
In contrast, the neural activity associated with visual imagery is more distributed and complex~\cite{holm2024contribution, orima2023spatiotemporal}. 
Given that our objective for the visual imagery task was to achieve object-level differentiation for fine-grained, real-world items, the obtained accuracy is a promising initial result. 


\begin{figure}[h] 
    \centering
    \vspace{-0.2in}
    \includegraphics[width=0.9\columnwidth]{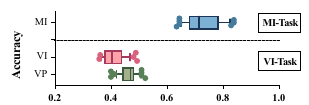} 
    \caption{\textbf{Comparison of decoding accuracies for VI and MI.} 
    Motor imagery achieves higher accuracy than the visual EEG tasks overall, and within the visual paradigm, VP slightly outperforms VI.
    \label{fig:offline_comapre}}
\vspace{-0.2in}
\end{figure}

\paragraph{Task Comparison}
As shown in Figure~\ref{fig:offline_comapre}, a comparison between the VI and MI tasks revealed significant differences in performance. 
For the VI task, decoding accuracy for perception was consistently higher than for imagery. This finding is in line with previous studies, suggesting that direct visual stimuli evoke a stronger and more distinct neural response compared to imagined visual stimuli~\cite{kilmarx2024evaluating, lee2022exploring}. 

While the accuracy for the VI task may appear modest, it is crucial to consider the specific context of our experimental setup. 
At the system level, this level of accuracy makes the task of grasping specific fruits using EEG signals feasible and effective.
Despite the lower direct visual decoding accuracy in the VI task, our results consistently surpass chance levels, indicating the potential of 
using EEG-based system to effectively support fruit-grasping tasks.

\paragraph{Average Evoked Responses}

\begin{figure}[htbp]
    \centering

    \begin{subfigure}[b]{0.34\textwidth}
        \centering
        \includegraphics[width=\linewidth]{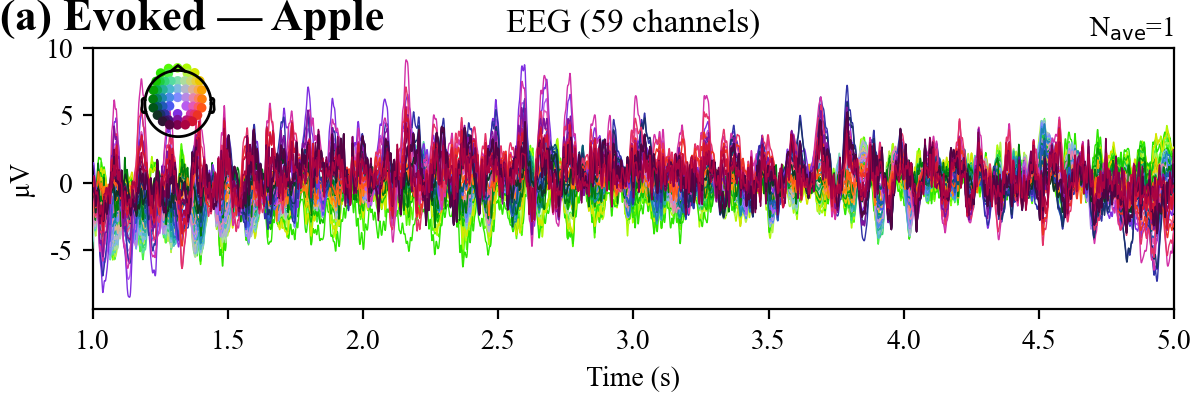}
    \end{subfigure}

    \begin{subfigure}[b]{0.34\textwidth}
        \centering
        \includegraphics[width=\linewidth]{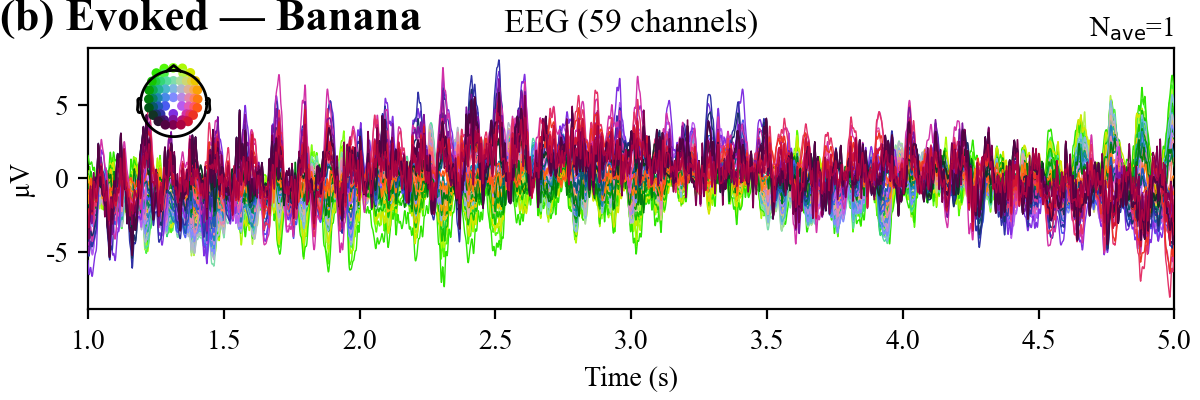}
    \end{subfigure}

    \begin{subfigure}[b]{0.34\textwidth}
        \centering
        \includegraphics[width=\linewidth]{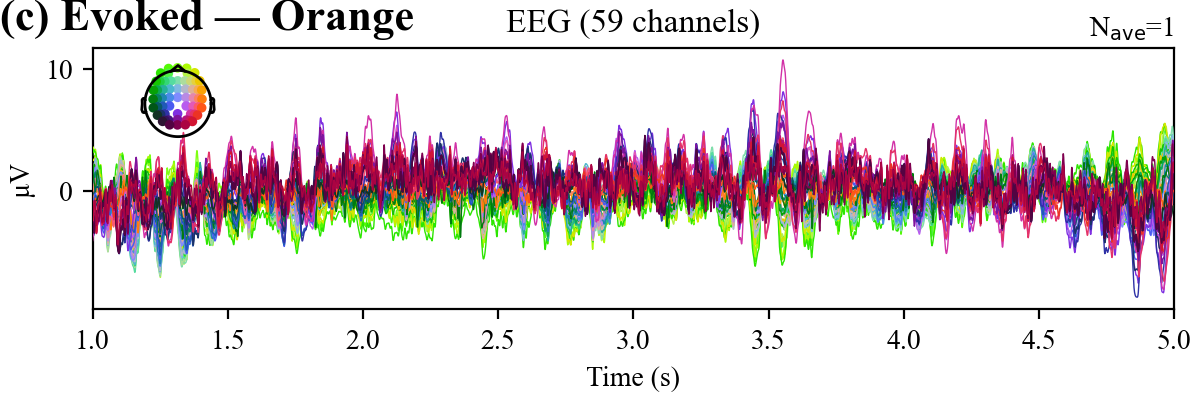}
    \end{subfigure}

    \caption{\textbf{Evoked EEG responses to fruit visual imagery.} 
    Brain activity patterns during VI for three stimuli in Sub00 (59 channels): (a) Apple, (b) Banana, (c) Orange. Responses vary over time yet remain broadly similar across categories, reflecting the fine-grained nature of VI decoding.
    }
    \label{fig:evoked}
    \vspace{-0.15in}
\end{figure}

Figure~\ref{fig:evoked} shows the average event-related potentials (ERPs) for Sub00 during the visual imagery task, illustrating how brain activity evolves in response to different stimuli. Brain activity showed clear temporal variation, with ERP differences across fruits indicating that visual imagery tasks can modulate neural responses to distinct stimuli and support classification.  However, despite this variation, a fairly consistent ERP pattern emerges across signals for all three fruit conditions. This uniformity of the brain's response across categories also indicates the inherent challenge in discriminating between conditions, emphasizing the complexity of interpreting EEG signals in this context.

\section{ONLINE DEMOSTRATION}

Based on the collected offline data and pre-trained model, we conducted online experiments for both VI and MI tasks. 
The system was deployed within a real-time data transmission pipeline with EEG streamed from the NeuroHUB platform to the robotic control system. 
All online experiments were conducted on a laptop running Ubuntu 20.04, equipped with an Intel i7-7700HQ CPU and an NVIDIA GeForce GTX 1070 GPU.

\subsection{Online System Workflow} 
In the \textbf{basic demonstration} scenario, the task was to recognize different fruits and place them in the correct positions. 
This demonstration highlights the system's ability to translate mental imagery into actionable robotic commands for the selected object and the intended position.
The online test for our base demonstration followed a structured, multi-stage workflow to ensure reliable human–robot interaction.

\textbf{\textit{Preparation and Task Initiation.}} 
Each trial began with a beep prompt. Participants performed an online-VI task for 15 seconds by mentally visualizing the target object, followed by an online-MI task for 15 seconds by imagining left or right hand movements, consistent with the offline protocol.

\textbf{\textit{Neural Decoding and Inference.}} Real-time EEG was streamed to pretrained VI and MI decoders. 
VI decoding identified the target object, while MI decoding determined the placement position, providing the robot with the information needed to compute the optimal interaction location.

\textbf{\textit{Action Execution.}} The system integrated VI (object) and MI (position) results to guide grasp-and-place actions.
Based on these inferences, the manipulator grasped the target object and placed it at the inferred location.

By converting neural intentions into robotic actions, the system effectively bridged mental imagery with physical execution, demonstrating reliable real-time interaction between human cognition and the robotic platform.

\subsection{System Performance and Results}

\begin{table*}[!t]
    \centering
    \caption{\textbf{Online system results.} \textbf{Top:} Base demo performance per subject on online VI/MI tasks across \( \text{model} / \text{frequency} \) settings \textit{(e.g., 63.64~(RGNN/40) denotes accuracy using the RGNN model at 40~Hz)}. For each subject, we report the top-2 online accuracies and the overall mean. \textbf{Bolded values} indicate the row maximum, while \underline{underlined values} \textbf{mark} the second highest. \textbf{Middle:} Overall online system accuracy, with online VI/MI accuracy representing the mean top-2 accuracies across all subjects. \textbf{Bottom:} Overall system runtime statistics.}
    \label{tab:online_results_and_times}

    \begin{minipage}{\textwidth}
    \centering

    \begin{tabularx}{\textwidth}{>{\centering\arraybackslash}p{0.6cm}*{6}{>{\centering\arraybackslash}X}}
        \toprule
        \multicolumn{7}{c}{\textbf{Online Base Demonstration results}} \\
        \cmidrule(lr){1-7}
        \multirow{2}{*}{\centering \textbf{Subject}} & 
        \multicolumn{3}{c}{\textbf{VI-online Task}} & 
        \multicolumn{3}{c}{\textbf{MI-online Task}} \\
        \cmidrule(lr){2-4} \cmidrule(lr){5-7}
        & \makecell{\textbf{Top1 Acc. (\%)}} & \makecell{\textbf{Top2 Acc. (\%)}} & \makecell{\textbf{Overall Avg. (\%)}} & \makecell{\textbf{Top1 Acc. (\%)}} & \makecell{\textbf{Top2 Acc. (\%)}} & \makecell{\textbf{Overall Avg. (\%)}} \\
        \cmidrule(lr){1-7}

        01 & \makecell{63.64~(RGNN/40)} & \makecell{44.44~(RGNN/60)} & \textbf{47.14} & \makecell{75.00~(RGNN/60)} & \makecell{60.00~(MLP/40)} & \textbf{61.67} \\
        02 & \makecell{50.00~(EEGNet/100)} & \makecell{39.13~(MLP/100)} & \underline{40.82} & \makecell{65.22~(MLP/100)} & \makecell{60.00~(MLP/40)} & 58.41 \\
        03 & \makecell{50.00~(MLP/60)} & \makecell{33.67~(RGNN/60)} & 38.89 & \makecell{60.00~(RGNN/100)} & \makecell{53.85~(RGNN/60)} & 54.62 \\
        04 & \makecell{37.50~(RGNN/100)} & 33.67~(MLP/60) & 34.72 & \makecell{66.67~(MLP/60)} & \makecell{60.00~(MLP/40)} & \underline{60.00} \\
        \midrule
    \end{tabularx}

    \begin{tabularx}{\textwidth}{*{5}{Y}}
        \multicolumn{5}{c}{\textbf{Online System Accuracy}} \\
        \cmidrule(lr){1-5}
        \textbf{Online VI avg Acc. (\%)} &
        \textbf{Query success rate (\%)} &
        \textbf{Online MI avg Acc. (\%)} &
        \textbf{Place success rate (\%)} &
        \textbf{System Accuracy (\%)} \\
        \cmidrule(lr){1-5}
        \textbf{40.23} & 76.11 & \textbf{62.59} & 100 & \textbf{20.88} \\
        \midrule
        \multicolumn{5}{c}{\textbf{System operation times (s)}} \\
    \end{tabularx}
    
    \begin{tabularx}{\textwidth}{*{9}{Y}}
        \cmidrule(lr){1-9}
        Prepare & VI Task & \mbox{VI Data Proc.} & VI Infer & MI Task & \mbox{MI Data Proc.} & MI Infer & Robot Exec. & System Total \\
        \cmidrule(lr){1-9}
        6.010 & 15.000 & 0.639 & \textbf{8.191} & 15.000 & 0.524 & \textbf{8.000} & \textbf{54.872} & \textbf{107.266} \\
        \bottomrule
    \end{tabularx}
    \end{minipage}
    \vspace{-0.2in}
\end{table*}

The online system results were evaluated with four participants in a base demonstration scenario. Each participant completed at least 50 iterations of the \textit{Online System Workflow} to validate the entire system. Following the offline accuracy hierarchy, we tested each model-frequency configuration (model/Hz), running at least 4-5 trials per configuration. A model was not considered effective for online application if its results showed no change. Models that performed best in the offline setting did not necessarily yield the same success rate in the online scenario, indicating that factors such as model stability and other conditions also play a role in real-time performance. After initial evaluation of all configurations, models with relatively higher success rates were selected for at least 15 online trials to assess the system's overall performance. 

Table~\ref{tab:online_results_and_times} presents the online system results in detail.

\paragraph{\textbf{Detailed Online Performance Analysis}}
Upon completing experiments with the four participants, we selected the top-2 online accuracies for both the VI and MI tasks, along with their corresponding \( \text{model} / \text{frequency} \) configurations. Among all participants, Sub01 demonstrated the best overall online performance, with average accuracies of \textbf{47.14\%} for the online-VI task and 
\textbf{61.67\%} for the online-MI task. 
Online inference used a 10-second window (3–13s) within the 15-second task. This temporal constraint, combined with less refined nature of real-time preprocessing, largely explains the reduced accuracies compared to offline results. 
For Sub03 and Sub04, certain VI conditions (model/Hz) exhibited systematic biases (e.g., apples misclassified as bananas, bananas as oranges) and were therefore excluded from the final analysis. 
Overall, the mean online accuracy across all subjects, including configurations not shown in the top-2, was 40.23\% for VI and 62.59\% for MI, reflecting the impact of inter-subject variability, physiological state, and the inherent characteristics of EEG signals in the online setting.
Notably, among the top-performing configurations, MLP/100 and RGNN/60 consistently performed well, suggesting promise for real-time use.

\paragraph{\textbf{Overall System and Operational Performance}} From the system perspective, the online system accuracy was \textbf{20.88\%} (Table~\ref{tab:online_results_and_times}, middle). 
This composite metric integrates visual recognition accuracy, motor position recognition accuracy, and the robotic arm’s object acquisition and placement success rates. While the overall score is modest, it indicates that the primary bottleneck is converting noisy EEG signals into reliable commands for robotic execution. Notably, the placement success rate was 100\%, showing that once the system correctly identified the placement position, execution was consistently successful. 
The average runtime per demonstration was \textbf{107.266 s} (Table~\ref{tab:online_results_and_times}, bottom). VI and MI inference take about 8\,s each per task, with average preprocessing times of 0.639\,s (VI) and 0.524\,s (MI), totaling about 17\,s for EEG decoding (\(\sim\)16\% of the runtime). In contrast, robot execution accounts for the largest portion (54.872\,s), indicating that timing is mainly constrained by system-level operations rather than the EEG pipeline.
Taken together, these findings indicate that strengthening EEG decoding robustness and optimizing task design are the most effective levers for overall reliability and performance.

\newcolumntype{Y}{>{\centering\arraybackslash}X}
\begin{table}[h]
\vspace{-0.1in}
    \centering
    \caption{Grasp Success Rates for Different Fruits}
    \label{tab:grasp_success_rate}
    \begin{tabularx}{\linewidth}{Y Y} 
        \toprule
        \textbf{Fruit Catalogue} & \textbf{Success Rate (\%)} \\
        \midrule
        Apple  & 89.83 \\
        Banana & 55.84 \\
        Orange & 84.44 \\
        \midrule
        \textbf{Overall Success Rate} & \textbf{76.11} \\
        \bottomrule
    \end{tabularx}
\vspace{-0.1in}
\end{table}

\paragraph{\textbf{Robotic Grasping and System Reliability}}

The average grasp query success rate was \textbf{76.11\%}, as shown in Table \ref{tab:grasp_success_rate}, which includes the success rates for different objects in the fruit catalogue. Specifically, apples and oranges had high success rates of \textbf{89.83\%} and \textbf{84.44\%}, respectively, while the success rate for bananas was comparatively lower at 55.84\%. 
These discrepancies can be attributed to small execution and pose estimation inaccuracies (e.g., a depth-related Z error of about $\pm$2\% for the RealSense D435 at a 2\,m range, and slight deviations in 6D grasp pose computation). Elongated, non-uniform objects such as bananas are more sensitive to such deviations. Despite these challenges, the system maintained a reasonable grasping performance overall.

\begin{figure}[htbp] 
    \centering
    \vspace{-0.1in}
    \includegraphics[width=1.0\columnwidth]{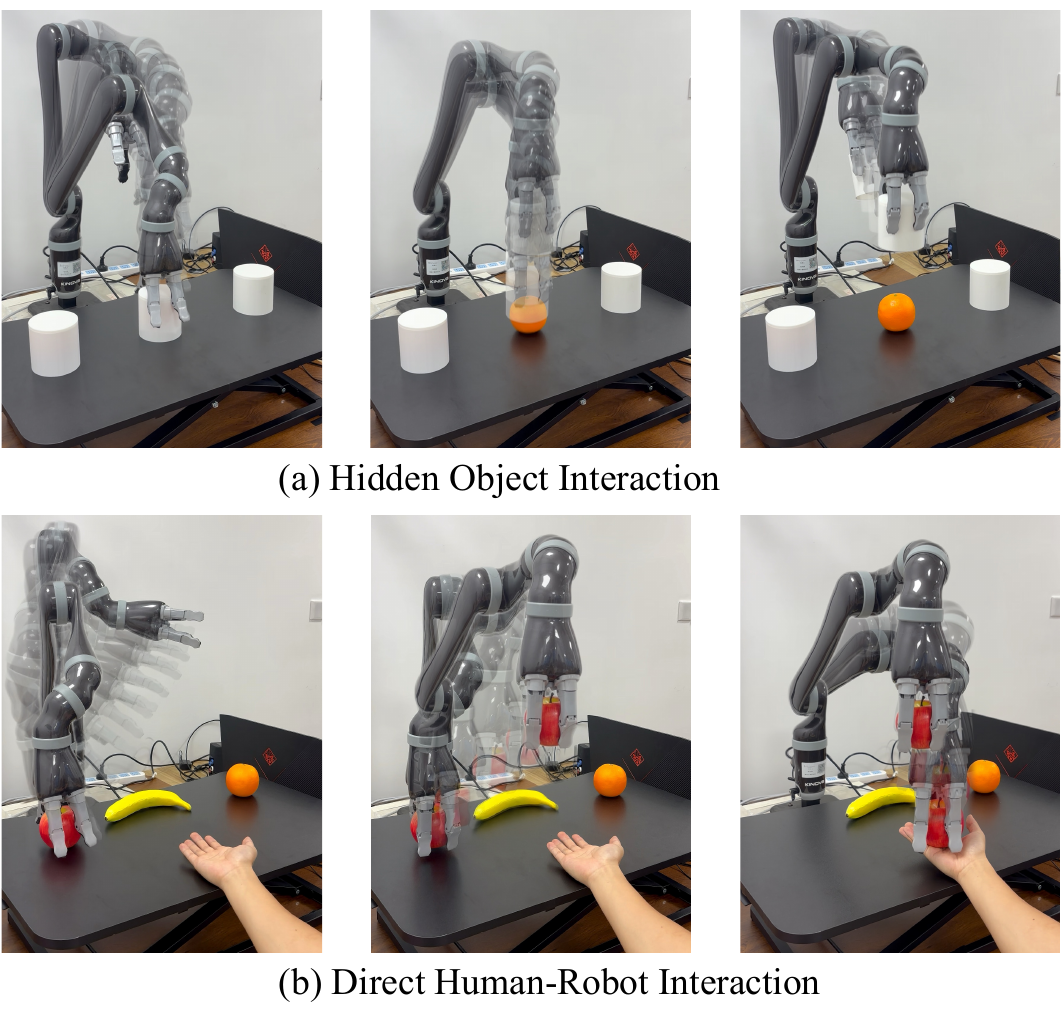} 
    \caption{\textbf{Online Demos for extended application scenarios.} \textbf{(a) Hidden Object Interaction:} Using only VI-EEG, the subject guides the system to identify and reveal a hidden object, revealing an ability to decode mental imagery without visual cues. \textbf{(b) Direct Human-Robot Interaction:} By imagining the object and controlling hand movements through Visual Imagery and Motor Imagery, the subject interacts seamlessly with the robot, which places the object directly into their hand. 
    \label{fig:example}}
    \vspace{-0.2in}
\end{figure}

\subsection{Extended Application Scenarios}

Beyond the base demonstration, we conducted a series of online tests in various scenarios to further validate the system's robustness and adaptability. \textbf{A key strength of our system is its posture-independent operation.} Tasks rely solely on mental imagery without additional visual or environmental cues, allowing the system to function in any environment or orientation. In addition to validating the system in a seated posture, \textbf{we also successfully completed online tests with subjects in a supine position,} as shown in the lower left corner of Figure~\ref{fig:framework}.

To further assess the system’s performance in environments with no visual input, we designed specific experiments to test its ability to decode neural signals and interact with objects. Demonstrations are presented in Figure~\ref{fig:example}.

\noindent \textbf{\textit{Hidden Object Interaction: EEG Decoding Without Visual Input.}} 
The \textbf{H}idden \textbf{O}bject \textbf{I}nteraction task was designed to evaluate the system’s ability to decode EEG-derived mental imagery and perform object manipulation without any visual information. In this experiment, various objects were placed on a table and concealed from the subject’s view with lids. Without knowing the type or location of the objects, the subject relied solely on mental imagery to guide the system. EEG signals corresponding to the subject’s visual imagination were decoded to identify the intended object, and the robot then manipulated and revealed the object that matched the subject’s mental imagery.

\noindent\textbf{\textit{Direct Human-Robot Interaction: Object Manipulation with Visual and Motor Imagery.}}
The \textbf{D}irect \textbf{H}uman-Robot \textbf{I}nteraction task aimed to evaluate the system’s ability to support seamless, intuitive interaction between humans and robots. In this scenario, the subject first visualized a target by VI, and the system decoded the EEG signals to identify the intended object. The subject then performed MI by imagining left- or right-hand movements while extending the corresponding hand. The system decoded these motor signals to determine the intended hand position, enabling the robot to grasp the target and place it directly into the subject’s hand. This task demonstrated the integration of visual and motor imagery for reliable, real-time human-robot interaction.


\section{CONCLUSIONS}

In this study, we present a high-level cognitive control paradigm for natural intention-driven BCI, where VI-EEG decodes grasp intentions and MI-EEG infers target placement positions. By combining offline pretraining with a real-time framework, we achieved seamless mapping from visual cognition to robotic actions, enabling accurate grasping and reliable placement of real-world objects in online scenarios. 
Although the system accuracy can still be improved, our results demonstrate the feasibility of leveraging higher-order cognitive states for intuitive brain–robot interaction. Current limitations mainly include the intrinsic constraints of EEG, insufficient feature separability of VI-EEG, and the need for improved long-term robustness and generalizability, which point to promising directions for future work.

\bibliographystyle{IEEEtran}
\bibliography{reference}

\end{document}